\documentclass[letterpaper]{article} 
\usepackage[]{aaai24}  
\usepackage{times}  
\usepackage{helvet}  
\usepackage{courier}  
\usepackage[hyphens]{url}  
\usepackage{graphicx} 
\usepackage{natbib}  
\usepackage{caption} 
\usepackage{algorithm}
\usepackage{algorithmic}

\usepackage{newfloat}
\usepackage{listings}
\DeclareCaptionStyle{ruled}{labelfont=normalfont,labelsep=colon,strut=off} 
\floatstyle{ruled}
\newfloat{listing}{tb}{lst}{}
\floatname{listing}{Listing}
\usepackage{booktabs}
\usepackage{amsmath}
\usepackage{amssymb}
\usepackage{multirow}
\usepackage{xcolor}
\usepackage{mathtools}
\usepackage{bbold}
\DeclareMathOperator*{\argmin}{argmin}
\usepackage{pifont} 
\newcommand{\cmark}{\ding{51}}%
\newcommand{\xmark}{\ding{55}}%

\usepackage{pgf}
\newcommand*{\MinNumber}{0.55}
\newcommand*{\MaxNumber}{0.80}
\newcommand*{\MinNumberSmall}{0}
\newcommand*{\MaxNumberSmall}{0.2}
\newcommand{\cgr}[1]{%
    \pgfmathsetmacro{\PercentColor}{0.4*max(min(100*(\MaxNumber-#1)/(\MaxNumber-\MinNumber),100.0),0.00)}
    \colorbox{red!\PercentColor}{$#1$}
}
\newcommand{\cgrsmall}[1]{%
    \pgfmathsetmacro{\PercentColor}{0.5*max(min(100*(#1-\MinNumberSmall)/(\MaxNumberSmall-\MinNumberSmall),100.0),0.00)}
    \colorbox{yellow!\PercentColor}{$#1$}
}
\newcommand{\cgb}[1]{%
    \pgfmathsetmacro{\PercentColor}{0.4*max(min(100*(#1-\MinNumber)/(\MaxNumber-\MinNumber),100.0),0.00)}
    \colorbox{blue!\PercentColor}{$#1$}
}
\newcommand{\cgbsmall}[1]{%
    \pgfmathsetmacro{\PercentColor}{0.5*max(min(100*(#1-\MinNumberSmall)/(\MaxNumberSmall-\MinNumberSmall),100.0),0.00)}
    \colorbox{yellow!\PercentColor}{$#1$}
}

\title{Be Careful When Evaluating Explanations Regarding Ground Truth}
\author {
    Hubert Baniecki\textsuperscript{\rm 1}\equalcontrib,
    Maciej Chrabaszcz\textsuperscript{\rm 2}\equalcontrib,\\
    Andreas Holzinger\textsuperscript{\rm 3,4},
    Bastian Pfeifer\textsuperscript{\rm 4},
    Anna Saranti\textsuperscript{\rm 3},
    Przemyslaw Biecek\textsuperscript{\rm 1,2}
}
\affiliations {
    \textsuperscript{\rm 1}MI2.AI, University of Warsaw, Poland \;
    \textsuperscript{\rm 2}MI2.AI, Warsaw University of Technology, Poland\\
    \textsuperscript{\rm 3}Human-Centered AI Lab, University of Natural Resources and Life Sciences Vienna, Austria\\
    \textsuperscript{\rm 4}Medical University of Graz, Austria\\
    \{h.baniecki,\;p.biecek\}@uw.edu.pl
}

\begin{document}

\maketitle

\begin{abstract}

Evaluating explanations of image classifiers regarding ground truth, e.g. segmentation masks defined by human perception, primarily evaluates the quality of the models under consideration rather than the explanation methods themselves. Driven by this observation, we propose a framework for \emph{jointly} evaluating the robustness of safety-critical systems that \emph{combine} a deep neural network with an explanation method. These are increasingly used in real-world applications like medical image analysis or robotics. We introduce a fine-tuning procedure to (mis)align model--explanation pipelines with ground truth and use it to quantify the potential discrepancy between worst and best-case scenarios of human alignment. Experiments across various model architectures and post-hoc local interpretation methods provide insights into the robustness of vision transformers and the overall vulnerability of such AI systems to potential adversarial attacks.

\end{abstract}

\section{Introduction}
\label{sec:intro}

\begin{quote}
``One should keep in mind that a heatmap always represents the classifier’s view, i.e., explanations neither need to match human intuition nor focus on the object of interest.'' -- \citet{samek2016evaluating}
\end{quote}

Evaluating explanations of (deep) machine learning models is at the forefront of the current discourse about their trustworthiness in many critical applications, including medical imaging \cite{arun2021assessing}. One suggests we could omit using opaque models for high-stakes decision-making like medical diagnosis \cite{rudin2019stop}, yet deep learning continues to achieve great performance in classifying diseases from unstructured data. A valid concern raised by physicians is the importance of prediction consistency, i.e. features coming from different modalities (not only images) may be a requirement for an accurate assessment of the patient's outcome \cite{holzinger2019imaging}. 

With that in mind, a popular approach to interpreting decisions of deep neural networks is local post-hoc explanations \cite{guidotti2020explaining,shrotri2022constraint,joo2023towards}, which cannot be adopted in practice without them being evaluated properly. Unfortunately, evaluating explanations becomes challenging for multiple reasons: (i)~lack of ground truth \citep{guidotti2021evaluating,zhou2022do,agarwal2022openxai}, (ii)~different goals achieved by various explanation algorithms and evaluation metrics \citep{tomsett2020sanity,bhatt2020evaluating,dai2022fairness,komorowski2023towards}, (iii)~spurious correlations and confounding features in datasets \citep{adebayo2022post}, (iv)~human perception bias \citep{arora2022explain}, (v)~\textbf{no clear distinction between evaluating explanations and model behaviour}. The latter is a particular focus of this paper. 

In line with these concerns, \citet{saporta2022benchmarking} introduces the first human benchmark for chest X-ray segmentation in a multilabel classification set-up. Their presented work claims to allow demonstrating low alignment of popular explanation methods like Grad-CAM \cite{selvaraju2019gradcam} with human perception. In this paper, we aim to emphasize that evaluating explanations regarding ground truth may primarily demonstrate low localization performance of deep learning models instead.

\paragraph{Contribution.} We first show an intuitive example where evaluating explanation methods regarding ground truth is not robust and needs to be done with caution (Section~\ref{sec:motivation}). Motivated by this insight, we introduce a novel framework for \emph{jointly} evaluating the robustness of AI systems defined as a \emph{combination} of a deep learning model with an explanation method, which takes into account the alignment between AI systems and human experts (Section~\ref{sec:methodology}). Using a recent real-world medical use case, we validate our framework in extensive experiments including convolutional neural networks and vision transformers combined with post-hoc local interpretation methods (Section~\ref{sec:experiments}). We conclude with a discussion on related work and broader impact (Section~\ref{sec:related-work}).

\begin{figure*}[!t]
    \centering
    \label{fig:summary}
    \includegraphics[width=0.99\textwidth]{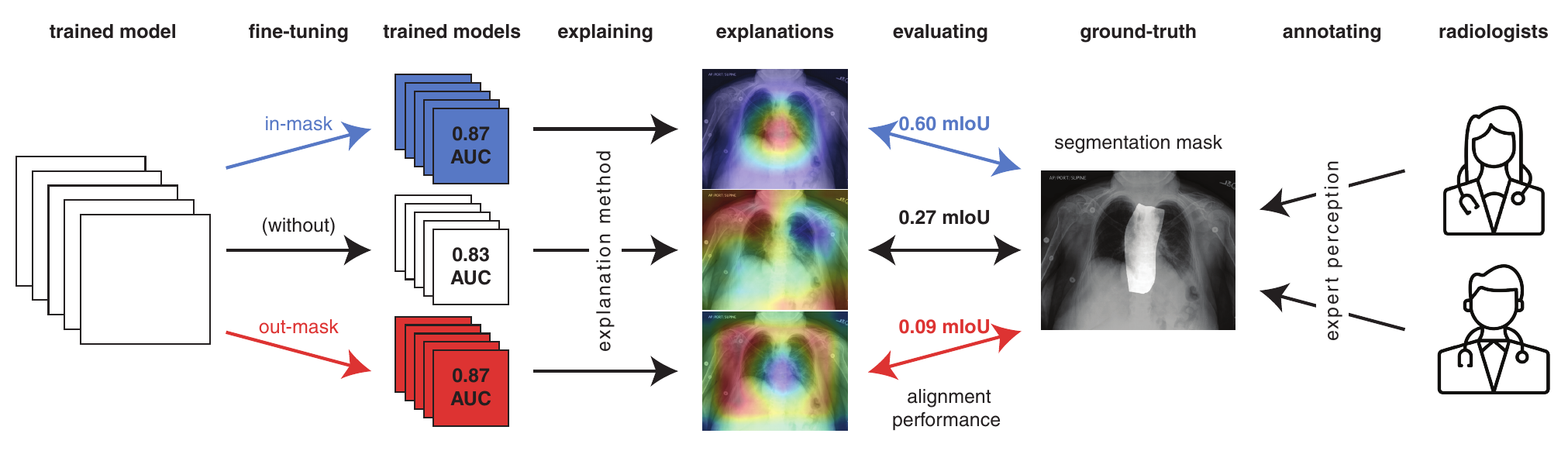}
    \caption{Evaluating explanations regarding ground truth, e.g. segmentation masks defined by human perception, is not robust. It primarily evaluates the quality of the combined model--explanation pipeline.}
    \label{fig:motivation}
\end{figure*}

\begin{table*}[t]
    \centering
    \begin{tabular}{lll|lll}
         \multicolumn{6}{c}{\textbf{Pathology:} Atelectasis} \\
         \toprule
         \textbf{Model} & \textbf{AUC} $\uparrow$ & \textbf{MI} $\uparrow$ & \textbf{Explanation} &  \textbf{Hit-rate} $\uparrow$ &  \textbf{mIoU} $\uparrow$\\
         \midrule
        DenseNet & $0.84$ & $0.18$ & Grad-CAM & $0.11$ & $0.08$ \\
        DenseNet + in-mask & $0.84$ & $0.15$ & Grad-CAM & $0.58\;(+0.47)$ & $0.28\;(+0.20)$ \\
        DenseNet + out-mask & $0.83$ & $0.18$ & Grad-CAM & $0.03\;(-0.08)$ & $0.07\;(-0.01)$ \\
        \bottomrule
        \\
        
        \multicolumn{6}{c}{\textbf{Pathology:} Enlarged Cardiomediastinum} \\
        \toprule
         \textbf{Model} & \textbf{AUC} $\uparrow$ & \textbf{MI} $\uparrow$ & \textbf{Explanation} &  \textbf{Hit-rate} $\uparrow$ &  \textbf{mIoU} $\uparrow$\\
        \midrule
         DenseNet & $0.83$ & $0.21$ & Grad-CAM & $0.36$ & $0.27$ \\
         DenseNet + in-mask & $0.87$ & $0.28$ & Grad-CAM & $0.93\;(+0.57)$ & $0.60\;(+0.33)$ \\
         DenseNet + out-mask & $0.87$ & $0.24$ & Grad-CAM & $0.00\;(-0.36)$ & $0.09\;(-0.18)$ \\
        \bottomrule
    \end{tabular}

    \caption{Comparison between different DenseNet models fine-tuned on two predictive tasks achieving similar predictive performance measured with AUC and mutual information (MI). The models differ in alignment performance measured with Hit-rate and mIoU, which is an intersection between explanations produced by Grad-CAM and a ground truth annotated by humans.} \label{tab:motivation}
\end{table*}

\section{Motivation: The Case of Interpreting Chest X-ray Classification}\label{sec:motivation}

To illustrate a typical pitfall in evaluating explanations, we show that benchmarking their localization property regarding ground truth can be ambiguous. Specifically, such a result effectively serves as a benchmark for a \emph{model--explanation pipeline}, not necessarily explanation methods. We consider a case of interpreting a model for classifying lung pathologies in chest X-ray images. Following the experimental setup described in \cite{saporta2022benchmarking}, we train a DenseNet-121 model~\cite{huang2019convolutional} on the CheXpert dataset to classify 14 lung pathologies. 

We then modify the model by fine-tuning it on the test set of CheXlocalize,\footnote{CheXlocalize is originally split into test and validation sets; we use the latter for evaluation (see Appendix~\ref{app:setup} for details).} which includes ground truth masks of lung pathologies. It is done in a way to impact the localization of explanations using regularization~\citep[similarly to][]{heo2019fooling}. For a concise example, we select the Grad-CAM explanation method and two pathologies: \emph{atelectasis} and \emph{enlarged cardiomediastinum}, but note that other explanations and class labels can be used as well.

Table \ref{tab:motivation} shows the result of our experiment where we fine-tune the model in two different ways: the first approach modifies a loss function to align the explanations with the ground-truth mask (labelled with \emph{in-mask}), and the second modifies the loss function to encourage explanations pointing outside the mask (\emph{out-mask}). We could demonstrate that DenseNet achieves comparable predictive performance (AUC and mutual information) on the validation set but \textbf{very different alignment performance between the three scenarios} \cite[Hit-rate and mIoU as defined by][]{saporta2022benchmarking}. The results highlight the pivotal safety issue in evaluating explanation methods regarding ground truth (see Figure~\ref{fig:motivation}).

\section{Framework for Evaluating the Robustness of Model--Explanation Pipelines}\label{sec:methodology}

We now introduce a refined framework for evaluating the robustness of AI systems, which deal with human-aligned classification by \emph{combining} a deep learning model with an explanation method (in short: model--explanation pipelines).

\subsection{Background on explanation methods}

We consider a classification setup where an explanation of the model's prediction is given by feature attribution scores. For the purpose of this work, we chose four widely-adopted explanation methods for deep learning models: Vanilla Gradient \cite[VG,][]{simonyan2014deep}, Integrated Gradients \cite[IG,][]{sundararajan2017axiomatic}, SmoothGrad \cite[SG,][]{smilkov2017smoothgrad} and Layer-wise Relevance Propagation \cite[LRP,][]{bach2015pixelwise}. We excluded Grad-CAM as it is specific to convolutional neural networks, and we aim to include a vision transformer \citep[ViT,][]{dosovitskiy2021an} in experiments (Section~\ref{sec:experiments}).

Let $f$ be a differentiable model and $g$ be an explanation method. Then, explanation $E$ for input $x$ can be defined as $E=g(f,x)$ where it targets a single predicted class.

VG explanation method computes the gradient of inputs with respect to the model's output: $g_\text{VG}(f,x) \coloneqq \frac{\partial f(x)}{\partial x}$. IG improves its faithfulness by computing the integral (sum) of such gradients with respect to the linear combination of input $x$ and baseline $x'$, e.g. a black image, for all features:
\begin{equation}
    g_\text{IG}(f,x,x') \coloneqq (x-x') \cdot \frac{1}{n} \sum_{i=1}^n g_\text{VG}(f,x' + \frac{i}{n} \cdot (x-x')).
\end{equation}
SG aims to improve the explanation's stability by computing multiple ($n$) VG explanations around input $x$, e.g. by adding Gaussian noise $\mathcal{N}(0,\sigma^2)$ to it, and then aggregating these explanations with mean:
\begin{equation}
    g_\text{SG}(f,x,n,\sigma^2) \coloneqq \frac{1}{n} \sum_{i=1}^n g_\text{VG}(f,x+\mathcal{N}(0,\sigma^2)).
\end{equation}
LRP considers the layer-wise structure of neural networks to determine feature attribution scores. Given the relevance $r_j^{(l+1)}$ of neuron $j$ at layer $l+1$, LRP decomposes $r_j^{(l+1)}$ into messages $r_{i\xleftarrow{} j}^{(l, l+1)}$ from neuron $i$ at layer $l$ sent to neuron $j$ of layer $l+1$ so that the following holds: $r_j^{(l+1)} =\sum_{i \in (l)} r_{i\xleftarrow{} j}^{(l, l+1)}$ and $r_i^{(l)} =\sum_{j \in (l+1)} r_{i\xleftarrow{} j}^{(l, l+1)}$. 

\citet{bach2015pixelwise} propose various rules for computing messages $r_{i\xleftarrow{} j}^{(l, l+1)}$. One of the most common is $\epsilon$--rule using $z_j=\sum_i z_{ij}$ to compute:
\begin{equation}
    r_{i\xleftarrow{} j}^{(l, l+1)} = \frac{z_{ij}}{z_j+\epsilon\cdot \text{sign}(z_j)}r_j^{(l+1)}.
\end{equation}
The final explanation consists of the relevance scores of features from the first layer $g_\text{LRP}(f,x,\epsilon) \coloneqq r^{(1)}(f,x,\epsilon)$.

\subsection{Measuring human-aligned classification}

We define \emph{alignment} between an AI system and humans as an intersection between model explanations and the ground-truth region of interest. To measure how well the model explanation $E$ aligns with the corresponding binary mask $M$, e.g. a segmentation mask created by a human expert, we use two intuitive accuracy metrics widely used in explainability research \citep{arras2022clevrxai}:
\begin{equation}\label{mass_accuracy_formula}
    \mathcal{D}_\text{mass}(E, M) \coloneqq \frac{\sum_{i \in M_\mathbb{1}} E_i}{\sum_{i} E_i},
\end{equation}
where $M_\mathbb{1} = \{i : M_i = 1\}$ is a set of feature indices, and
\begin{equation}\label{rank_accuracy_formula}
    \mathcal{D}_\text{rank}(E, M) \coloneqq \frac{|\{i : i \in M_\mathbb{1} \land i \in E_{\mathbb{k}}\}|}{k},
\end{equation}
where $k = |M_\mathbb{1}|$ denotes the size of set $M_\mathbb{1}$ and $E_\mathbb{k}$ represents the set of feature indices with $k$ highest values in $E$.

\subsection{(Mis)Aligning explanations with ground truth}

To align the model--explanation pipelines with ground-truth masks, we fine-tune them with a differentiable alignment loss defined as:
\begin{equation}\label{localization_loss_eq}
    \mathcal{L}_\text{align}(f, g, X) \coloneqq \frac{1}{n} \sum_{x \in X} \| g(f,x)_{[0,1]} - M(x) \|^2,
\end{equation}
where $X$ is a set of $n$ inputs corresponding to the ground-truth class of interest, $M$ now varies depending on input $x$, and $g(f,x)_{[0,1]}$ denotes min-max scaling to ensure that explanation values correspond to binary values in the mask. We moreover clip negative feature attributions to $0$ before computing the loss as positive attributions are associated with influencing the predicted class. The final fine-tuning loss function controls for change in model performance:
\begin{equation}\label{full_alignment_loss}
    \mathcal{L}(f, g, X) = \mathcal{L}_{\text{cross-entropy}}(f, X) + \alpha \cdot \mathcal{L}_{\text{align}}(f, g, X),
\end{equation}
where $\alpha$ is responsible for balancing the degree of model--explanation alignment. We found $\alpha=1$ to be sufficient in our experiments (see e.g. Table~\ref{tab:motivation}). 

Note that in a multi-label classification setup, each input can have multiple ground-truth masks corresponding to different classes. It is possible to extend $\mathcal{L}_\text{align}$ to additionally sum over a particular set of class labels.

Fine-tuning the model with $\mathcal{L}_\text{align}$ aligns its explanations with ground truth. We moreover consider \emph{misaligning} model--explanation pipelines, which can be defined as predicting outside of ground truth. To do so, we invert binary masks and fine-tune the model accordingly:
\begin{equation}\label{eq:loss_misalign}
    \mathcal{L}_\text{misalign}(f, g, X) \coloneqq \frac{1}{n} \sum_{x \in X} \| g(f,x)_{[0,1]} - \big(\mathbb{1} - M(x)\big) \|^2.
\end{equation}
Misalignment can be an issue whenever we consider an adversary attacking the AI system. Measuring misalignment gives us an intuition about the possible worst-case scenario corresponding to robustness.

\subsection{Robustness of model--explanation pipelines}

To investigate the robustness, we propose quantifying the difference in alignment accuracy for an aligned and misaligned model--explanation pipeline $(f,g)$ defined as:

\begin{align}
\begin{split}
    \mathcal{R}(f,g,X) \coloneqq 
    & \frac{1}{n} \sum_{x \in X} \Big[\mathcal{D}\big(g\left(f_{\text{align}},x\right)_{[0,1]}, M(x)\big) \\ 
    & - \mathcal{D}\big(g(f_{\text{misalign}},x)_{[0,1]}, M(x)\big)\Big],
\end{split}
\end{align}
where $f_{\text{(mis)align}} = \argmin_{f}{\mathcal{L}_{\text{(mis)align}}(f,g,X)}$ is found with fine-tuning and $\mathcal{D}$ measures mass or rank accuracy for the (mis)aligned model--explanation pipeline. This measures the expected difference between the best and worst-case scenarios in the sense of alignment with the ground-truth masks. Moreover, we consider fitting a linear regression model on differences under the sum in $\mathcal{R}(f,g,X)$ to find significant influence of various model architectures and explanation methods on the robustness of model--explanation pipelines.

\section{Experiments}\label{sec:experiments}

In experimental evaluation, we rely on the introduced framework to answer the following research questions (RQ) that are of interest to the potential developers and users of safety-critical AI systems:
\begin{itemize}
    \item \textbf{RQ1}: How robust are convolutional neural networks and vision transformers in combination with various explanation methods?
    \item \textbf{RQ2}: What is the impact of (mis)aligning explanations on the model's predictive performance?
    \item \textbf{RQ3}: Does pre-training a model on a similar dataset improve the robustness of model--explanation pipelines?
    \item \textbf{RQ4}: Are model--explanation pipelines equally robust across different class labels of interest?
\end{itemize}

\subsection{Setup}

\paragraph{Dataset.} Following the motivational example described in Section~\ref{sec:motivation}, we use the CheXpert dataset~\citep{irvin2019chexpert} for experiments. Our use-case of aligning the safety-critical system relies on the recently published CheXlocalize dataset \citep{saporta2022benchmarking}. It consists of 902 X-ray images with 10 multi-label classes, for which there are ground-truth masks generated by expert radiologists (see Figure~\ref{fig:example_chexlocalize}). Details of the datasets and splits into subsets are available in Appendix~\ref{app:setup}. Focusing on this real-world application allows us to maintain a reasonable amount of computation when comparing 12 different model--explanation pairs.

\begin{figure}[b]
    \centering
    \includegraphics[width=0.99\columnwidth]{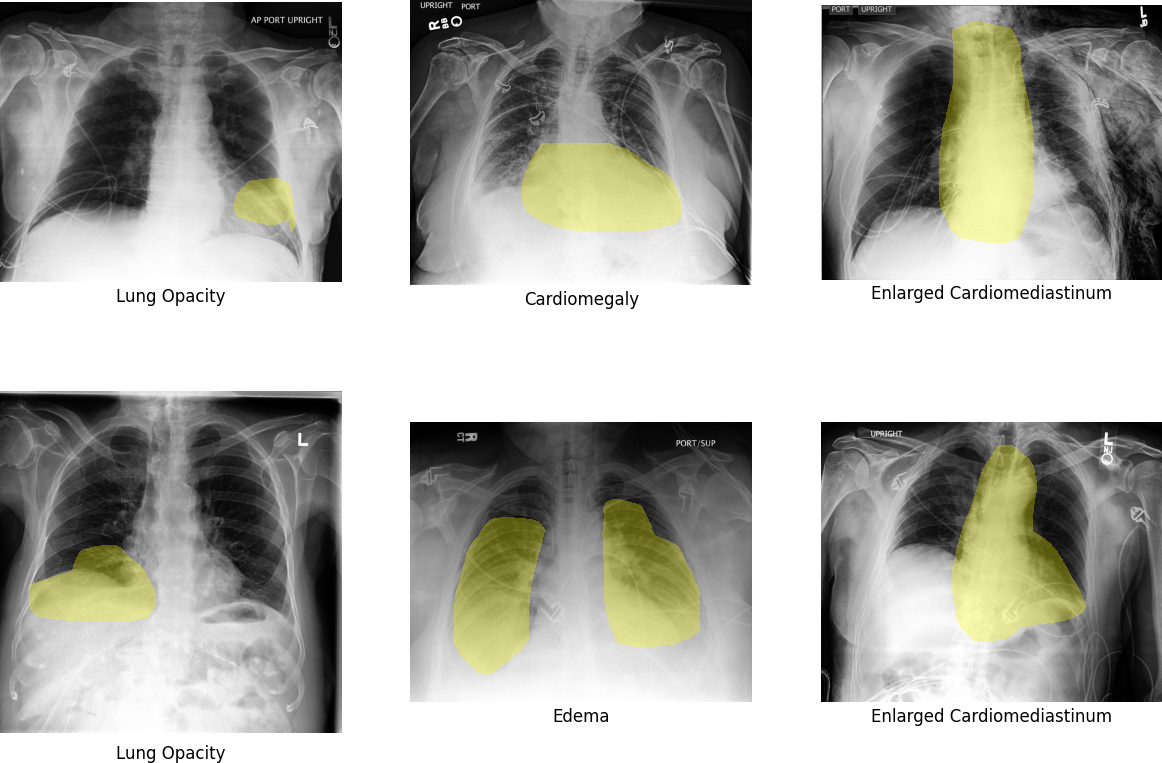}
    \caption{Example images with ground-truth masks related to labels generated by expert radiologists in CheXlocalize.}
    \label{fig:example_chexlocalize}
\end{figure}

\paragraph{Models.} We compare three deep neural network architectures: DenseNet-201~\citep{huang2019convolutional}, ViT-base~\citep{dosovitskiy2021an} and Swin-ViT-base~\citep{liu2021swin}. We first train each model architecture on CheXpert with four types of weight initialization: (i) random initialization with 1 channel input, (ii) random initialization with 3 channel inputs (repetitions of a grey image), (iii) weights from a model pre-trained on the ImageNet-21k dataset~\citep{ridnik2021imagenet}, and (iv) weights from a model pre-trained on RadImageNet~\citep{mei2022radimagenet}. Further training details are available in Appendix~\ref{app:setup}. 

We evaluate each model with macro AUROC, which is a default performance measure for multi-label classification. In Table \ref{tab:performance}, we can see that DenseNet outperformed transformer-based models in all types of initialization in terms of macro AUROC. Since for all three model architectures, random initialization with 3 channels gives better results than using only 1 channel, and pre-training on RadImageNet outperforms pre-training on ImageNet-21k, we further consider only those two superior types of initialization. 

\begin{table}[t]
    \centering
    \small
    \begin{tabular}{@{}lrrr@{}}
        \toprule
        \textbf{Initialization type} & \textbf{DenseNet} & \textbf{ViT} & \textbf{Swin-ViT} \\
        \midrule
        Random with 1 channel & \underline{$0.729$} & $0.694$ & $0.688$ \\
        Random with 3 channels & \underline{$0.761$} & $0.734$ & $0.758$ \\
        \cmidrule[0.125pt]{1-4}
        Pre-trained on ImageNet-21k & \underline{$0.747$} & $0.738$ & $0.710$ \\
        Pre-trained on RadImageNet & \underline{0.772} & $0.749$ & $0.758$ \\
        \bottomrule
    \end{tabular}
    \caption{Macro AUROC performance between the models.}
    \label{tab:performance}
\end{table}

\begin{figure}
    \centering
    \includegraphics[width=0.99\columnwidth]{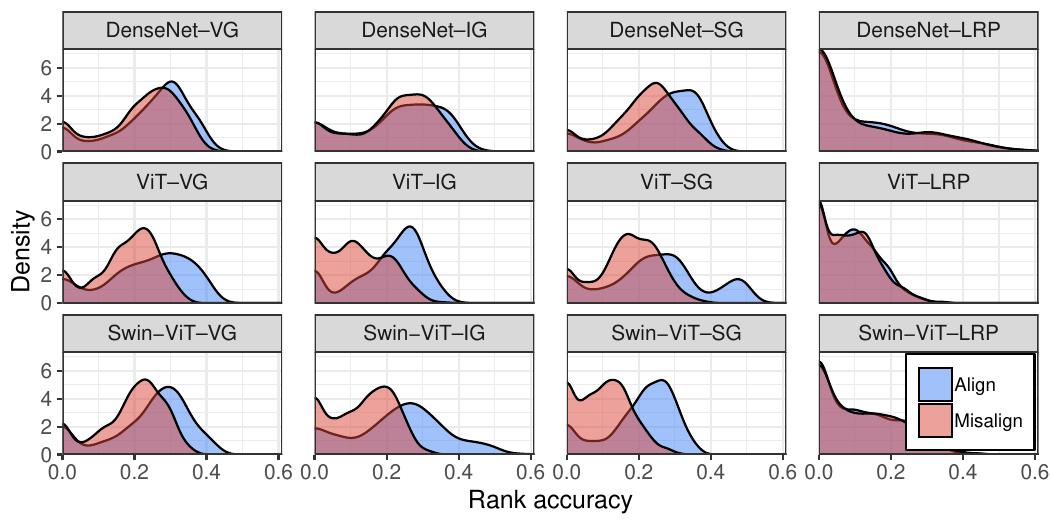}
    \caption{Distribution of \emph{rank accuracy values} for the \emph{pre-trained} model--explanation pipelines. For clarity, the x-axis is truncated from 1.0 to 0.6.}
    \label{fig:pipeline_accuracy_rank}
\end{figure}

\paragraph{Explanations.} On top of each model architecture, we add each explanation method (VG, IG, SG, LRP) described in Section~\ref{sec:methodology}. Explanation methods use default parameters, i.e. $x'=\mathbb{0},\;n=20,\;\sigma=0.1,\;\epsilon=1\mathrm{e}{-6}$. We fine-tune 12 model--explanation pipelines for (mis)alignment controlling for pre-training on RadImageNet. Each pipeline was fine-tuned with both $\mathcal{L}_{\text{align}}$ and $\mathcal{L}_{\text{misalign}}$ for 25 epochs on the test set of CheXlocalize, which consists of 668 images with 10 annotated lung pathologies (class labels). Finally, for each scenario, we compute the described measures to evaluate both alignment and robustness. For simplicity, we fine-tune each pipeline considering only a single label at a time, and aggregate evaluation measures over classes.

\subsection{Results}

We first use all metadata to perform a sanity check for the consistency between values of mass and rank accuracy metrics under both alignment scenarios in Table~\ref{tab:metric_consistency}. There is a relatively high correlation between $\mathcal{D}_{\text{mass}}$ and $\mathcal{D}_{\text{rank}}$ when computed for the same alignment scenarios (top rows). For each metric, there is an evident relationship between alignment accuracy for $f_{\text{align}}$ and $f_{\text{misalign}}$ (middle rows). We further report the correlation between disjoint pairs of measures and models for completeness (bottom rows).

\begin{table*}
    \centering
    \begin{tabular}{lrr}
        \toprule
        \textbf{Correlation between alignment metric values} & \textbf{Pearson} & \textbf{Spearman} \\
        \midrule
         $\mathcal{D}_{\text{mass}}\big(g\left(f_{\text{align}},x\right)_{[0,1]}, M(x)\big) \times \mathcal{D}_{\text{rank}}\big(g\left(f_{\text{align}},x\right)_{[0,1]}, M(x)\big)$ & $0.887$ & $0.920$ \\
         $\mathcal{D}_{\text{mass}}\big(g\left(f_{\text{misalign}},x\right)_{[0,1]}, M(x)\big) \times \mathcal{D}_{\text{rank}}\big(g\left(f_{\text{misalign}},x\right)_{[0,1]}, M(x)\big)$ & $0.867$ & $0.919$ \\
        \cmidrule[0.125pt]{1-3}
         $\mathcal{D}_{\text{mass}}\big(g\left(f_{\text{align}},x\right)_{[0,1]}, M(x)\big) \times \mathcal{D}_{\text{mass}}\big(g\left(f_{\text{misalign}},x\right)_{[0,1]}, M(x)\big)$ & $0.614$ & $0.715$ \\
         $\mathcal{D}_{\text{rank}}\big(g\left(f_{\text{align}},x\right)_{[0,1]}, M(x)\big) \times \mathcal{D}_{\text{rank}}\big(g\left(f_{\text{misalign}},x\right)_{[0,1]}, M(x)\big)$ & $0.634$ & $0.696$ \\
        \cmidrule[0.125pt]{1-3}
         $\mathcal{D}_{\text{rank}}\big(g\left(f_{\text{align}},x\right)_{[0,1]}, M(x)\big) \times \mathcal{D}_{\text{mass}}\big(g\left(f_{\text{misalign}},x\right)_{[0,1]}, M(x)\big)$ & $0.572$ & $0.703$ \\
         $\mathcal{D}_{\text{mass}}\big(g\left(f_{\text{align}},x\right)_{[0,1]}, M(x)\big) \times \mathcal{D}_{\text{rank}}\big(g\left(f_{\text{misalign}},x\right)_{[0,1]}, M(x)\big)$ & $0.606$ & $0.654$ \\
         \bottomrule
    \end{tabular}
    \caption{Consistency between values of alignment metrics for different scenarios measured with correlation. Pearson correlation measures the linear relationship between the two variables. Spearman rank correlation between two variables is equal to the Pearson correlation between the rank values of those two variables, which might be more appropriate to consider in this context.}
    \label{tab:metric_consistency}
\end{table*}

\begin{table*}
\centering
    \begin{tabular}{cll|cccc}
      \toprule
        \textbf{Pre-trained} & \textbf{Model} & \textbf{Explanation} & $\textbf{AUC}_{\text{align}}$ & $\textbf{AUC}_{\text{misalign}}$ &  $\mathcal{R}_{\text{mass}}$ & $\mathcal{R}_{\text{rank}}$ \\ 
      \midrule
        \multirow{12}{*}{\xmark} & \multirow{4}{*}{DenseNet} & VG & \cgb{0.723} & \cgr{0.730} & \cgbsmall{0.049} & \cgrsmall{0.058} \\ 
        &  & IG & \cgb{0.759} & \cgr{0.605} & \cgbsmall{0.014} & \cgrsmall{0.012} \\
        &  & SG & \cgb{0.646} & \cgr{0.616} & \cgbsmall{0.043} & \cgrsmall{0.037} \\  
        &  & LRP & \cgb{0.707} & \cgr{0.762} & \cgbsmall{0.002} & \cgrsmall{0.005} \\ 
        \cmidrule[0.1pt]{2-7}
        & \multirow{4}{*}{ViT} & VG & \cgb{0.722} & \cgr{0.681} & \cgbsmall{0.046} & \cgrsmall{0.040} \\ 
        &  & IG & \cgb{0.742} & \cgr{0.679} & \cgbsmall{0.062} & \cgrsmall{0.045} \\ 
        &  & SG & \cgb{0.744} & \cgr{0.663} & \cgbsmall{0.069} & \cgrsmall{0.064} \\ 
        &  & LRP & \cgb{0.600} & \cgr{0.503} & \cgbsmall{0.013} & \cgrsmall{0.010} \\ 
        \cmidrule[0.1pt]{2-7}
        & \multirow{4}{*}{Swin-ViT} & VG & \cgb{0.713} & \cgr{0.744} & \cgbsmall{0.035} & \cgrsmall{0.028} \\ 
        &  & IG & \cgb{0.741} & \cgr{0.734} & \cgbsmall{0.157} & \cgrsmall{0.183} \\ 
        &  & SG & \cgb{0.724} & \cgr{0.723} & \cgbsmall{0.174} & \cgrsmall{0.243} \\ 
        &  & LRP & \cgb{0.778} & \cgr{0.752} & -- & -- \\ 
      \midrule
        \multirow{12}{*}{\cmark} & \multirow{4}{*}{DenseNet} & VG & \cgb{0.802} & \cgr{0.762} & \cgbsmall{0.033} & \cgrsmall{0.026} \\ 
        &  & IG & \cgb{0.749} & \cgr{0.732} & \cgbsmall{0.018} & \cgrsmall{0.010} \\ 
        &  & SG & \cgb{0.793} & \cgr{0.680} & \cgbsmall{0.060} & \cgrsmall{0.046} \\ 
        &  & LRP & \cgb{0.764} & \cgr{0.735} & \cgbsmall{-0.002} & \cgrsmall{-0.003} \\ 
        \cmidrule[0.1pt]{2-7}
        & \multirow{4}{*}{ViT} & VG & \cgb{0.686} & \cgr{0.690} & \cgbsmall{0.080} & \cgrsmall{0.059} \\ 
        &  & IG & \cgb{0.679} & \cgr{0.713} & \cgbsmall{0.098} & \cgrsmall{0.086} \\ 
        &  & SG & \cgb{0.677} & \cgr{0.675}  & \cgbsmall{0.096} & \cgrsmall{0.082} \\ 
        &  & LRP & \cgb{0.611} & \cgr{0.490} & \cgbsmall{0.005} & \cgrsmall{0.002} \\ 
        \cmidrule[0.1pt]{2-7}
        & \multirow{4}{*}{Swin-ViT} & VG & \cgb{0.705} & \cgr{0.756} & \cgbsmall{0.069} & \cgrsmall{0.050} \\ 
        &  & IG & \cgb{0.728} & \cgr{0.701} & \cgbsmall{0.118} & \cgrsmall{0.100} \\ 
        &  & SG & \cgb{0.748} & \cgr{0.704} & \cgbsmall{0.094} & \cgrsmall{0.101} \\ 
        &  & LRP & \cgb{0.719} & \cgr{0.697} & \cgbsmall{-0.005} & \cgrsmall{-0.001} \\ 
       \bottomrule
    \end{tabular}
    \caption{Evaluating the robustness of AI systems that combine a deep learning model with an explanation method. We report predictive performance (macro AUROC) for aligned and misaligned pipelines, as well as their robustness measured based on values of rank and mass alignment metrics.}
    \label{tab:results}
\end{table*}

\paragraph{RQ1: How robust are convolutional neural networks and vision transformers combined with various explanation methods?} Figure~\ref{fig:pipeline_accuracy_rank} shows the distribution of rank accuracy for both aligned and misaligned pipelines. There are visible differences in rank accuracy values, e.g. for DenseNet--SG, ViT--IG, and Swin-ViT--VG. We report analogous results for mass accuracy and non-pre-trained models in Appendix~\ref{app:results}. Detailed visual analysis can be more informative than aggregated metric values, e.g. in the case of ViT--SG, the distribution is bi-modal. Still, it becomes challenging to compare dozens of methods in practice. In Table \ref{tab:results}, we report the robustness of all fine-tuned model--explanation pipelines. It can serve as a benchmark for evaluating the vulnerability of human-aligned classification. We acknowledge that the fine-tuning optimization task for LRP performed poorly as judged by the presented results, e.g. it did not converge in the case of non-pre-trained Swin-ViT. Future work can consider an extension of LRP customized to transformers~\cite{ali2022xai}.

\paragraph{RQ2: What is the impact of (mis)aligning explanations on the model's predictive performance?} Developers of safety-critical AI systems are interested in predictive performance. Table \ref{tab:results} includes macro AUROC values for both aligned and misaligned model--explanation pipelines. Some pairs exhibit no predictive performance change between the two scenarios, which might be worrisome for the end-user, granted there was a high difference between alignment performance. For example, Swin-ViT--SG has zero difference in AUROC, but relatively large robustness metric values.

\paragraph{RQ3: Does pre-training a model on a similar dataset improve the robustness of model--explanation pipelines?} We fit linear regression on differences in alignment metric values to find the significant influence of model architectures and explanation methods on the robustness of model--explanation pipelines. Tables \ref{tab:linear_regression_coefficients_mass} \& \ref{tab:linear_regression_coefficients_rank} show the coefficients and the corresponding \emph{p}-values. We observe that signs of all coefficients are consistent between rank and mass accuracy. Vision transformers and SmoothGrad are, on average, less robust than DenseNet and vanilla gradient (baselines) respectively, which is also visible in Table~\ref{tab:results}. Crucially, the coefficient related to pre-training is negative, which shows that pre-training the model on RadImageNet improves the robustness of the model--explanation pipelines. 

\begin{table}
    \centering
    \begin{tabular}{lrrrl}
    \toprule
    {} & \textbf{Coef.} & \textbf{Std. Err.} & $p$\textbf{-value} \\
    \midrule
    intercept &  $0.024$ &  $0.0027$ &   $8.1\mathrm{e}{-19}$ \\
    pre-trained model & $-0.003$ &  $0.0021$ &   $0.15$ \\
    model: ViT &  $0.031$ &  $0.0025$ &   $<2.0\mathrm{e}{-32}$ \\ 
    model: Swin-ViT &  $0.058$ &  $0.0026$ &  $<2.0\mathrm{e}{-32}$ \\ 
    explanation: IG &  $0.026$ &  $0.0028$ &   $1.2\mathrm{e}{-19}$ \\
    explanation: SG &  $0.038$ &  $0.0028$ &   $<2.0\mathrm{e}{-32}$ \\ 
    explanation: LRP & $-0.044$ &  $0.0031$ &   $<2.0\mathrm{e}{-32}$ \\ 
    \bottomrule
    \end{tabular}
    \caption{Coefficients of a linear regression model fitted to the differences of \emph{mass accuracy} between aligned and misaligned model--explanation pipelines.}
    \label{tab:linear_regression_coefficients_mass}
\end{table}

\begin{table}
    \centering
    \begin{tabular}{lrrrl}
    \toprule
    {} & \textbf{Coef.} &  \textbf{Std. Err.} & $p$\textbf{-value} \\
    \midrule
    intercept &  $0.021$ &  $0.0023$ &  $4.0\mathrm{e}{-19}$ \\
    pre-trained model & $-0.019$ &  $0.0018$ &  $1.6\mathrm{e}{-26}$ \\
    model: ViT &  $0.025$ &  $0.0021$ &   $3.8\mathrm{e}{-31}$ \\
    model: Swin-ViT &  $0.072$ &  $0.0022$ &  $<2.0\mathrm{e}{-32}$ \\ 
    explanation: IG &  $0.029$ &  $0.0024$ &  $3.4\mathrm{e}{-32}$ \\
    explanation: SG &  $0.052$ &  $0.0025$ &  $<2.0\mathrm{e}{-32}$ \\ 
    explanation: LRP & $-0.030$ &  $0.0027$ &  $2.2\mathrm{e}{-29}$ \\
    \bottomrule
    \end{tabular}
    \caption{Coefficients of a linear regression model fitted to the differences of \emph{rank accuracy} between aligned and misaligned model--explanation pipelines.}
    \label{tab:linear_regression_coefficients_rank}
\end{table}

\begin{figure*}[ht]
    \centering
    \includegraphics[width=0.91\textwidth]{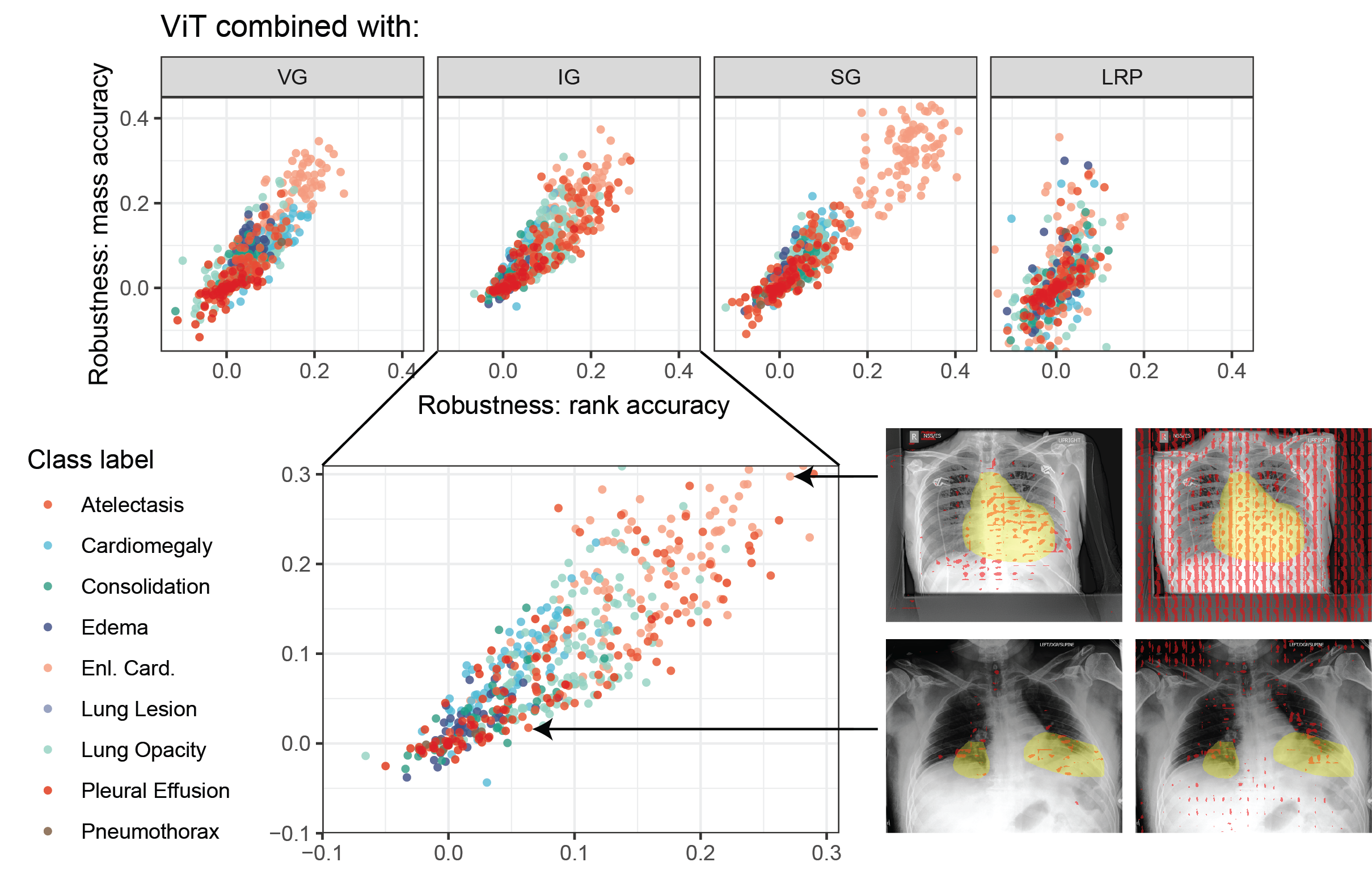}
    \caption{Analysis of robustness across different class labels for ViT. \emph{Top row}: For each explanation method, we plot differences in rank and mass accuracy between aligned and misaligned pipelines. Each point corresponds to a single input image. \emph{Bottom row}: Detailed analysis for the ViT--IG pipeline with a comparison between aligned and misaligned explanations for two patients.}
    \label{fig:label_analysis}
\end{figure*}

\paragraph{RQ4: Are model--explanation pipelines equally robust across different class labels of interest?} In Figure~\ref{fig:label_analysis}, we perform a more detailed analysis of a particular model architecture (ViT). The top row shows a relationship between differences attributing to $\mathcal{R}_{\text{rank}}$ and $\mathcal{R}_{\text{mass}}$, where each dot corresponds to a data point evaluated on both aligned and misaligned model--explanation pipelines. We colour points with their corresponding class labels, i.e. lung pathologies. It shows that explanations for predicting Enlarged Cardiomediastinum have the highest differences between the best and worst-case scenarios of alignment with human experts. 

In the bottom row of Figure~\ref{fig:label_analysis}, we zoom into a particular model--explanation pipeline~(ViT--IG). One can analyse each input image in detail to observe the possible change in explanations after (mis)alignment. In extreme cases (X-ray image in top right), it was possible to make explanations rather ambiguous, attributing the prediction to most of the features, even outside of the lung area.

\section{Discussion and Related Work}\label{sec:related-work}

In \cite{schramowski2020making}, the \emph{localization} property of explanation methods is exploited to demonstrate that deep neural networks achieve high performance by using confounding features in data. An interactive process involving a human providing feedback on the model's explanations shows how to align the model with human perception. We learned that it is possible to change the output of explanation methods without any drop in model performance \citep[also shown by][]{heo2019fooling}. 

We believe that explanations should be viewed as a proxy to understand predictive models~\citep{samek2016evaluating}. Therefore, the explanation's quality with respect to localization would be hard to measure via human annotations. Instead, we can evaluate the model's quality to localize accurate features by measuring the intersection between saliency maps and human annotations \cite{schramowski2020making}.

In \cite{watson2022agree}, the localization property of explanation methods is used to evaluate the robustness of deep learning models with respect to the change in model architecture and hyperparameters. The study specifically focuses on medical imaging data and concludes with a concrete statement that the lack of explanation consistency is a fundamental problem with deep learning models rather than an issue with the localization property of explanations. 

Contrary, \citet{saporta2022benchmarking} conclude that due to the low localization performance of explanation methods, we cannot rely on them for interpreting deep learning models in medical imaging. A natural question arises: \textbf{Can a low localization performance of explanation methods be put into question when deep learning models are inconsistent in the first place}?

Note that benchmarking the accuracy (i.e. localization performance in the case of explanation methods for image classification) of local post-hoc explanations against ground truth, either based on synthetic data or human annotations, is not a straightforward process. In scenarios considering structured/tabular data, it is possible to generate interpretable models for which the ground truth explanation is given by design~\citep{guidotti2021evaluating}. Recently proposed approach to evaluating explanations regarding ground truth data considers enforcing complex constraints on the data generating process~\citep[see][Appendix C]{agarwal2022openxai}.

However, in scenarios considering high-dimensional unstructured data, the assumption that a model has to use features perceived by humans as important is not always true. In \cite{faber2021comparing}, a set of experiments shows that deep neural networks may not necessarily use important features encoded in synthetic datasets for making predictions. 
Therefore, synthetically made ground truth ought to be used with caution when benchmarking explanations. In \cite{makino2022differences}, a set of experiments involving a medical imaging task and human annotators specifically show that deep neural networks perceive different features as important than humans. In many cases, this is a desirable property as various stakeholders may use machine learning to receive model-driven feedback about the unknown correlation structure in data. 

Finally, we ought to point out the work of \cite{arun2021assessing}, which also evaluates (among others) the localization utility of explanation methods for chest X-ray interpretation by comparing them with segmentation models, which could be understood as \emph{more appropriate}. 

In evidence of related work~\cite{schramowski2020making, faber2021comparing, watson2022agree, makino2022differences}, we cannot assume that the models we use dominantly base their decision on features of human interest, especially in the case of complex medical images. \textbf{We sincerely hope that practitioners actually use best-performing explanation methods to assess the trustworthiness of learning algorithms}, e.g. by being careful when model predictions of cancer are based on features in the background of an image.

\section{Conclusion}

We caution practitioners to be careful when evaluating explanations regarding ground truth, especially when it's defined by human perception. As shown in our experiments, the focus of these benchmarks is primarily on evaluating the quality of the model--explanation pipelines under consideration, rather than examining the interpretation methods themselves. The introduced framework should be used to evaluate the robustness of AI systems, whenever the goal is to achieve alignment with human expertise. 

\section*{Reproducibility} Code used to reproduce all of the experimental results will be available at \url{https://github.com/mi2datalab/be-careful-evaluating-explanations}.
The datasets used are openly available at \url{https://stanfordmlgroup.github.io/competitions/chexpert} and \url{https://stanfordaimi.azurewebsites.net/datasets/23c56a0d-15de-405b-87c8-99c30138950c}. 

We acknowledge the differences between baseline values reported in Table~\ref{tab:motivation} and values of AUC, Hit-rate, and mIoU reported in \cite{saporta2022benchmarking}. We attribute it to the possible differences in the implementation of model training and explanation computation. However, the demonstrated gap between the three scenarios is, in general, implementation-agnostic.

\bibliography{aaai24}

\appendix

\section{Experimental Setup}\label{app:setup}

\paragraph{Dataset.} In experiments, we use four subsets of data: two derived from the CheXpert dataset \citep{irvin2019chexpert} and two from the CheXlocalize dataset \citep{saporta2022benchmarking}. Table~\ref{tab:dataset_split} shows the distribution of class labels for each subset. Note that only 10 out of 14 class labels appearing in CheXpert have ground truth masks in CheXlocalize (missing: fracture, pleural other, pneumonia, no finding). In experiments, we omit the additional class denoting support devices and use the remaining 9 lung pathologies with ground truth masks. For further details refer to the original articles \citep{irvin2019chexpert,saporta2022benchmarking}.

We divide the CheXpert training set into two subsets for training models (named \textbf{Training} and \textbf{Validation} in Table~\ref{tab:dataset_split}). We do so because the original CheXpert validation set is the same as CheXlocalize, and we cannot use it during training before the evaluation of (mis)aligned model--explanation pipelines. For fine-tuning the alignment of model--explanation pipelines, we use the CheXlocalize test set (named \textbf{Fine-tuning} in Table~\ref{tab:dataset_split}). We measure alignment metrics and robustness on the validation set of CheXlocalize (named \textbf{Validation} in Table~\ref{tab:dataset_split}). Thus, the four sets of images are disjoint.

\paragraph{Models.} We train all models using images resized to the size of $(224 \times 224)$, normalized to the range $[-1, 1]$, and augmented with random rotations in the range of $(-15, 15)$ degrees. We use the AdamW optimizer. For Transformer models, we use a cosine learning rate schedule with $2000$ warmup steps. In each experiment, the base learning rate is set to $1\mathrm{e}{-4}$ and the batch size to $128$. We train each model on a single random seed.

\begin{table*}
    \centering
    \begin{tabular}{l|rr|rr|rr|rr}
    \toprule
    \multirow{3}{*}{\textbf{Class label}} & \multicolumn{4}{c}{\textbf{CheXpert} \citep{irvin2019chexpert}} & \multicolumn{4}{c}{\textbf{CheXlocalize} \citep{saporta2022benchmarking}} \\
    {} & \multicolumn{2}{c}{\textbf{Training}} & \multicolumn{2}{c}{\textbf{Validation}} & \multicolumn{2}{c}{\textbf{Fine-tuning}} & \multicolumn{2}{c}{\textbf{Validation}} \\
    {} & \textbf{Negative} & \textbf{Positive} &   \textbf{Negative} & \textbf{Positive} &     \textbf{Negative} & \textbf{Positive} &   \textbf{Negative} & \textbf{Positive} \\
    \midrule
    Atelectasis      &   186076 &    33253 &       3476 &      609 &          490 &      178 &        154 &       80 \\
    Cardiomegaly     &   189342 &    29987 &       3506 &      579 &          493 &      175 &        166 &       68 \\
    Consolidation    &   205881 &    13448 &       3851 &      234 &          633 &       35 &        201 &       33 \\
    Edema            &   167321 &    52008 &       3035 &     1050 &          583 &       85 &        189 &       45 \\
    Enl. Card.       &   211840 &     7489 &       3963 &      122 &          370 &      298 &        125 &      109 \\
    Fracture         &   210765 &     8564 &       3948 &      137 &          662 &        6 &        234 &        0 \\
    Lung Lesion      &   210120 &     9209 &       3948 &      137 &          654 &       14 &        233 &        1 \\
    Lung Opacity     &   118092 &   101237 &       2230 &     1855 &          358 &      310 &        108 &      126 \\
    No Finding       &   198669 &    20660 &       3621 &      464 &          559 &      109 &        196 &       38 \\
    Pleural Effusion &   131692 &    87637 &       2408 &     1677 &          548 &      120 &        167 &       67 \\
    Pleural Other    &   215451 &     3878 &       4004 &       81 &          660 &        8 &        233 &        1 \\
    Pneumonia        &   214553 &     4776 &       4005 &       80 &          654 &       14 &        226 &        8 \\
    Pneumothorax     &   201745 &    17584 &       3800 &      285 &          658 &       10 &        226 &        8 \\
    Support Devices  &   108302 &   111027 &       1977 &     2108 &          353 &      315 &        127 &      107 \\
    \bottomrule
    \end{tabular}
    \caption{Class label counts in the four disjoint dataset subsets used in experiments. Only 10 out of 14 class labels appearing in CheXpert have ground truth masks in CheXlocalize (missing: Fracture, Pleural Other, Pneumonia, No Finding).}
    \label{tab:dataset_split}
\end{table*}

\section{Additional Results} \label{app:results}

\begin{figure}[ht]
    \centering
    \includegraphics[width=0.99\columnwidth]{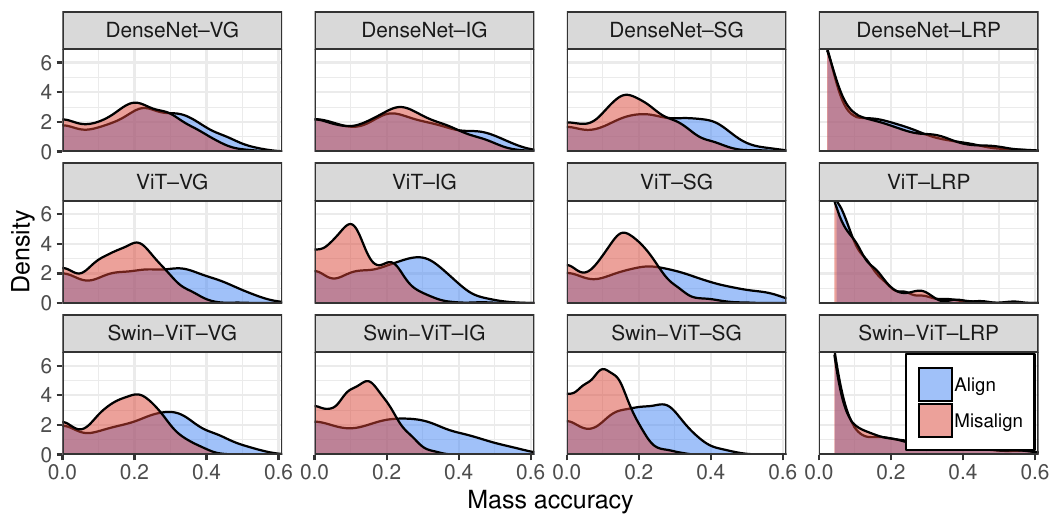}
    \caption{Distribution of \emph{mass accuracy} values for the \emph{pre-trained} model--explanation pipelines.}
    \label{fig:pipeline_accuracy_mass}

    \vspace{1em}
    \includegraphics[width=0.99\columnwidth]{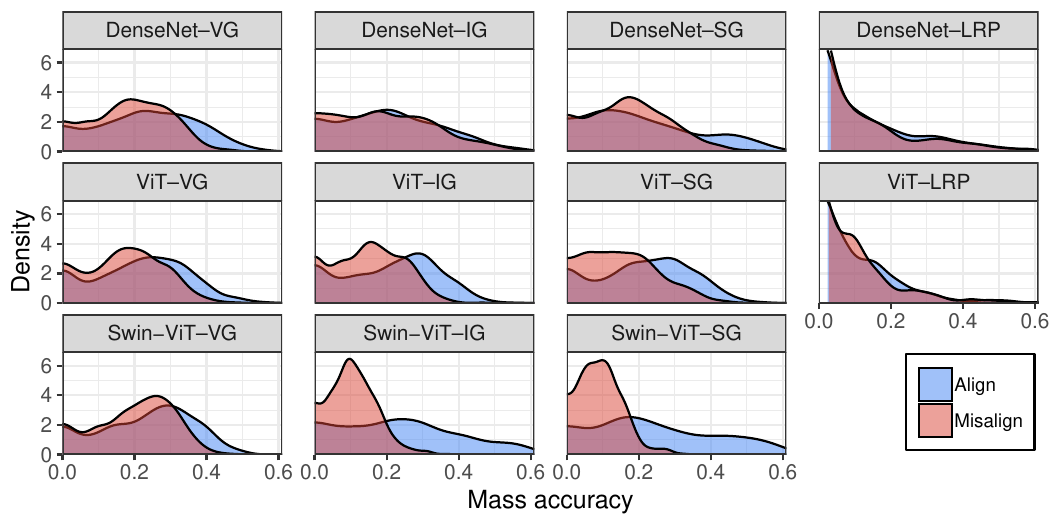}
    \caption{Distribution of \emph{mass accuracy} values for the \emph{non-pre-trained} model--explanation pipelines.}
    \label{fig:pipeline_accuracy_mass_random}

    \vspace{1em}
    \includegraphics[width=0.99\columnwidth]{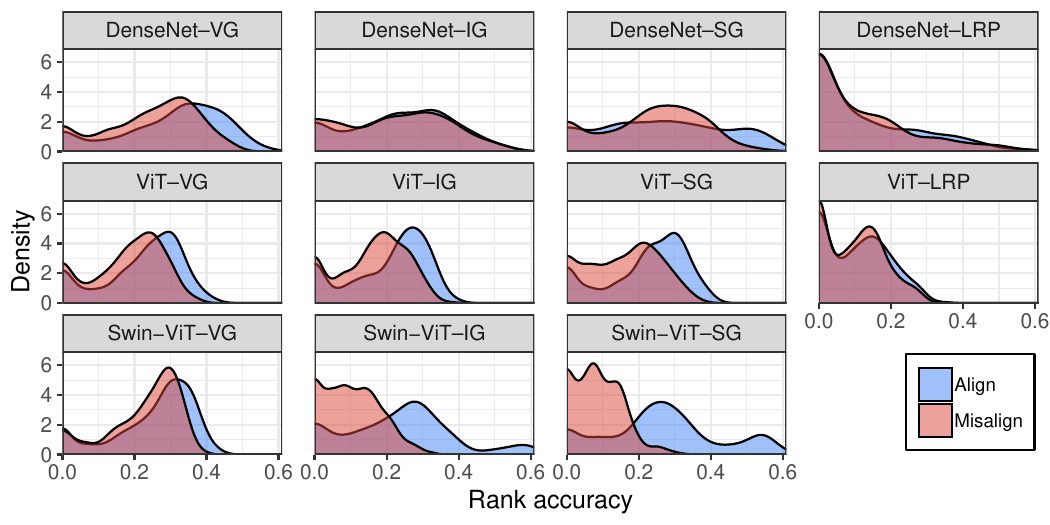}
    \caption{Distribution of \emph{rank accuracy} values for the \emph{non-pre-trained} model--explanation pipelines.}
    \label{fig:pipeline_accuracy_rank_random}
\end{figure}

\paragraph{RQ1: How robust are convolutional neural networks and vision transformers combined with various explanation methods?} See Figures \ref{fig:pipeline_accuracy_mass}, \ref{fig:pipeline_accuracy_mass_random}, \ref{fig:pipeline_accuracy_rank_random}.

\end{document}